# Brain Abnormality Detection by Deep Convolutional Neural Network

Mina Rezaei
Hasso Plattner Institute
14482-Prof.Dr.Helmer str.2
Potsdam-Germany
+49 331 5509-538
Mina.Rezaei@hpi.de

Haojin Yang
Hasso Plattner Institute
14482-Prof.Dr.Helmer str.2
Potsdam-Germany
+49 331 5509-511
Haojin.Yang@hpi.de

Christoph meinel
Hasso Plattner Institute
14482-Prof.Dr.Helmer str.2
Potsdam-Germany
T+49 331 5509-222
Christoph.meinel@hpi.de

## ABSTRACT
In this paper, we describe our method for classification of brain magnetic resonance (MR) images into different abnormalities and healthy classes based on deep neural network. We propose our method to detect high and low grade glioma, multiple sclerosis, and Alzheimer diseases as well as healthy cases. Our network architecture has ten learning layers that include seven convolutional layers and three fully connected layers. We have achieved promising result on five categories of brain images (classification task) with 95.7% accuracy.

## 1. INTRODUCTION
During the past years deep learning has raised a huge attention by showing promising results in some state-of-the-art approaches such as speech recognition, handwritten character recognition, image classification[1,2], detection[8][11] and segmentation [3,4]. There are expectations that deep learning improve or create medical image analysis applications, such as computer aided diagnosis, image registration and multimodal image analysis, image segmentation and retrieval. There has been some application that using deep learning in medical application like cell tracking [9] and organ cancer detection [10]. Doctors use Magnetic Resonance Imaging (MRI) as an effective tool to diagnose diseases. Medical application diagnosis tools make it faster and more accurate. The brain is a particularly complex structure, and analysis of brain magnetic resonance images is an important step for many problems such as surgical or chemical planning. As noted by Menze et al. [5] the number of publications involving automatic tumor detection has grown dramatically in the last decade. Previous work on human brain classification has achieved promising results via machine learning and classification techniques such as artificial neural networks and support vector machine (SVM) [6, 7] [13, 14].

In this work, we propose an architectures to the problem of brain abnormality detection. Our network is based on Krizhevsky et al. [1] with certain improvements. Our approach is end-to-end (single stage training and testing) with seven convolutional layers and three fully connected layers. In addition we consider data augmentation like randomly cropping, multiple scaling, horizontal and vertical mirroring and also drop out layer to prevent overfitting. Current network has two types of pooling layers (average and maximum) which reduce the dimension of the feature map by merging group of neurons. By putting support vector machine as loss function we have achieved better result.

## 2. Data Description
In this research we have used five different brain dataset to evaluate our proposed method.

### 2.1 Healthy Brain Images[1]
This data has collected nearly 600 MR images from normal, healthy subjects. The MR image acquisition protocol for each subject includes:
- T1, T2 and PD-weighted images
- MRA images
- Diffusion-weighted images in 15 directions

The data has been collected at three different hospitals in London and as part of IXI – Information extraction from Images project. The format of images is NIFTI (*.nii) and it is open access. Figure 1 column (a) shows healthy brain from IXI dataset in sagittal, coronal and axial section.

### 2.2 High and Low grade glioma (Tumor)[2]
This data is from BRATS (Brain Tumor Segmentation) challenges in MICCAI conference 2015. The data prepared in two parts training and testing for high and low grade glioma tumor. All data sets have been aligned to the same anatomical template and interpolated to 1mm^3 voxel resolution. The training dataset contains about 300 high- and low- grade glioma cases. Each dataset has T1 (spin-lattice relaxation), T1 contrast-enhanced MRI, T2 (spin-spin relaxation), and FLAIR MRI volumes. In the test dataset 200 MR images without label but in the same format is available. Figure1 column (b, c) shows high and low grade glioma, both categories are *.mha format.

### 2.3 Alzheimer disease[3]
The Alzheimer data set downloaded from Open Access Series of Imaging Studies (OASIS). The dataset consists of a cross-sectional collection of 416 subjects aged 18 to 96. For each subject, 3 or 4 individual T1-weighted MRI scans obtained in single scan sessions are included. The data format is *.hdr.

### 2.4 Multiple sclerosis[4]
Multiple Sclerosis MR Images downloaded from ISBI conference 2008 (The MS Lesion Segmentation Challenges).

---

[1] http://brain-development.org/ixi-dataset/

[2] https://www.virtualskeleton.ch/BRATS/Start2015/

[3] http://www.oasis-brains.org/

[4] http://www.medinfo.cs.ucy.ac.cy/index.php/downloads/datasets/

This data set collected by e-Health lab of Cyprus University. Figure 1 column (e) shows multiple sclerosis images which training dataset consists of 18 multiple sclerosis as *.nhdr format that ground truth (manual segmentation by expert) is available.

## 3.1 Training and Testing on GPU

We used Caffe [12] framework for training and testing. Our experiments were conducted on Ubuntu14.04, 16G RAM, and single Titan X GPU with 12G memory. Model has trained with learning rate η =0.001, weight decay λ= 0.0005. We trained model with 550k images for training with 20k iteration (we consider balancing sample from five categories in the test and train) and 157K in the testing with 5k iteration. Input images for our network are in RGB format. We convert and resize different formats of images to .png as input format for the caffe framework on VOI in three slices axial, coronal and sagittal. Then we randomly selected 70% of current balanced data for the training and 30% for testing.

## 4. Experiments and Result

For evaluating proposed method we computed accuracy, specificity and sensitivity. Table 1 describe our result on 1500 MR images.

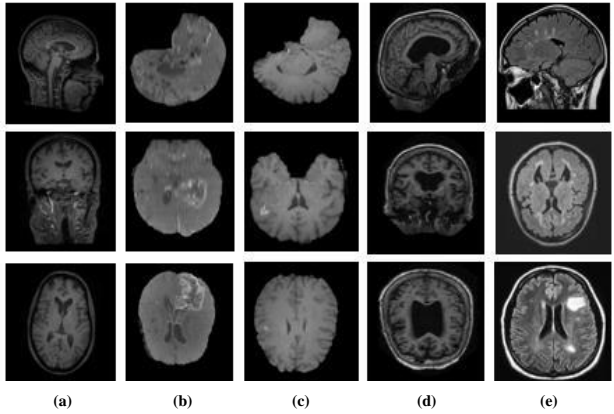

**Figure 1. We trained our architecture by five different categories of brain MRI. Column (a) shows healthy brain in sagittal, coronal and axial section. Column (b) and (c) show tumor high and low grad glioma. Column (d) and (e) present some brain data on Alzheimer and multiple sclerosis**

## 3. The Deep Architecture

Our network architecture for classification is based on the network proposed by Krizhevsky et al [1]. The architecture is summarized in Figure 2. Input image for the network has three channel and we use axial, coronal and sagittal slices in a Volume-of-Interest (VOI) for each categories. In order to prevent overfitting in the next step we exploit data augmentation. We increase convolutional layer to seven to achieve better results because it is deeper. We also consider computation complexity and the time of process by training and testing using very efficient GPU implementation of the convolution operation.

The network includes three pooling layers, average pooling after $3^{th}$ convolution layer and two max pool layer consequently after $5^{th}$, $6^{th}$ CNN layers that has effective impact on decreasing size of the feature map. At the end of 7th convolutional layer we put three fully-connected layers which have 4096 individual neurons. We apply regularization after last fully connected layer to reduce chance of overfitting. We put in the final a 5-way SVM to classify 4 abnormalities and healthy brain images.

|  | Total MRI | accuracy | sensitivity | specificity |
|---|---|---|---|---|
| Our propose method | 1500 | 95.07% | 0.91 | 0.87 |

**Table 1. Classification result on five different MR images by deep learning**

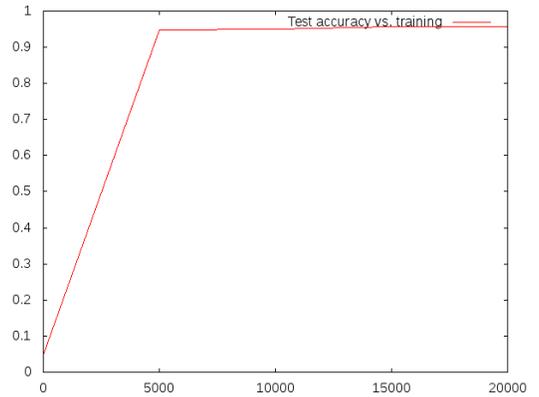

**Figure 3: Test accuracy vs training iterations**

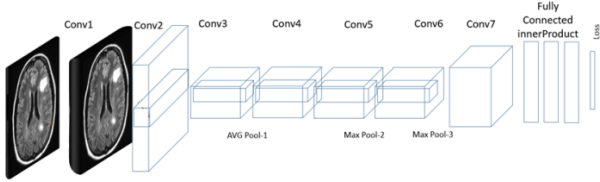

**Figure 2. Our convolution neural network consists of several convolutional layers, max and average pooling layers, fully-connected layers, a dropout layer, and a final 5-way SVM layer for classification.**

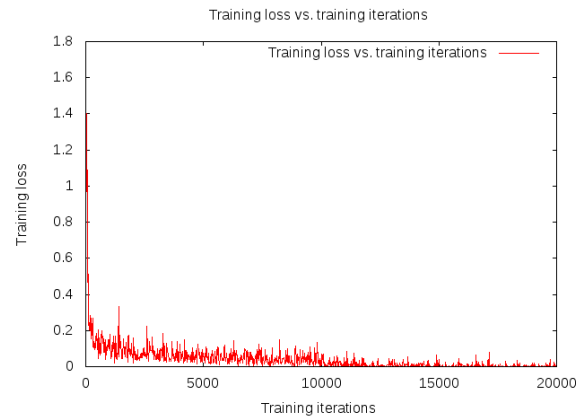

**Figure 4. Training loss according 2k iteration**

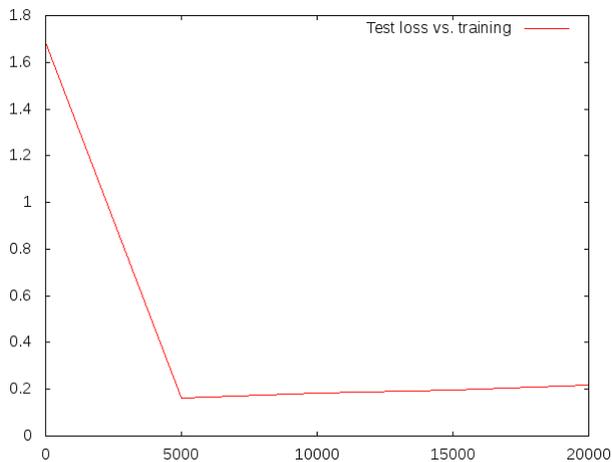

**Figure 5.** Training Iterations following Test Loss

## 5. Conclusion

We train our architecture on five different categories brain MR images. We have achieved promising results in automatic brain abnormality classification by using deep learning with 95.7% accuracy.